\documentclass[12pt,twoside]{article}
\usepackage{latexsym,amssymb}

\newenvironment{keywords}{\centerline{\bf\small
Keywords}\vspace{-2ex}\begin{quote}\small}{\par\end{quote}\vskip 1ex}

\newtheorem{nummer}{\hspace*{-0.33em}}[section]

\newenvironment{Def} {\begin{nummer} {\bf Definition.} \begin{rm}} {\end{rm} \end{nummer}}
\newenvironment{Lemma} {\begin{nummer} {\bf Lemma.}} {\end{nummer}}
\newenvironment{Prop} {\begin{nummer} {\bf Proposition.}} {\end{nummer}}
\newenvironment{Theorem} {\begin{nummer} {\bf Theorem.}} {\end{nummer}}
\newenvironment{Cor} {\begin{nummer} {\bf Corollary.}} {\end{nummer}}

\newenvironment{Problem} {\begin{nummer} {\bf Problem.}} {\end{nummer}}
\newenvironment{Example} {\begin{nummer} {\bf Example.} \begin{rm}} {\end{rm} \end{nummer}}
\newenvironment{MyPar}[1] {\begin{nummer} {\bf #1.} \begin{rm}} {\end{rm} \end{nummer}}

\newenvironment{Proof} {\vspace{1ex}\noindent \bf Proof. \rm}
  {\ \nolinebreak \hfill $\Box$ \vspace{2ex}}

\def\beq{\begin{equation}}
\def\eeq{\end{equation}}
\def\beqn{\begin{displaymath}}
\def\eeqn{\end{displaymath}}
\def\bqa{\begin{eqnarray}}
\def\eqa{\end{eqnarray}}
\def\bqan{\begin{eqnarray*}}
\def\eqan{\end{eqnarray*}}

\def\CC{\mathcal C}
\def\MM{\mathcal M}

\def\XX{\mathcal X}

\def\NNN{\mathbb N}
\def\RRR{\mathbb R}
\def\QQQ{\mathbb Q}

\def\BBB{\mathbb B}

\def\M{M}
\def\m{m}

\def\Km{K\!m}

\def\E{{\bf E}}

\def\rrho{\varrho}
\def\eps{\varepsilon}
\def\epstr{\epsilon}

\def\eqm{\stackrel\times=}
\def\leqm{\stackrel\times\leq}
\def\geqm{\stackrel\times\geq}
\def\neqm{\not\stackrel\times=}

\def\leqt{_{1:t}}
\def\ltt{_{<t}}
\def\leqn{_{1:n}}

\def\ltinf{_{<\infty}}
\def\_norm{_{norm}}

\newcommand{\zwidths}[1]{\rlap{$\scriptstyle #1$}}

\def\for_all{{\rm\ for\ all\ }}
\def\such_that{{\rm\ such\ that\ }}
\def\wenn{\mbox{\ if\ }}
\def\und{\mbox{\ and\ }}

\def\lb{{\log_2}}                      
\def\l{\ell}

\def\ph{\varphi}
\def\th{\vartheta}

\def\toims{\stackrel{i.m.s.}{\longrightarrow}}

\def\qmbox#1{{\quad\mbox{#1}\quad}}

\begin{document}

\title{\vskip -10mm\normalsize\sc Technical Report \hfill IDSIA-03-04
\vskip 2mm\bf\LARGE\hrule height5pt \vskip 5mm
\sc Convergence of Discrete MDL \\ for Sequential Prediction
\vskip 6mm \hrule height2pt \vskip 4mm}
\author{{\bf Jan Poland} and {\bf Marcus Hutter} \\[3mm]
\normalsize IDSIA, Galleria 2, CH-6928\ Manno-Lugano,
  Switzerland\thanks{This work was supported by SNF grant 2100-67712.02.}\\
\normalsize \{jan,marcus\}@idsia.ch, \ http://www.idsia.ch/$^{_{_\sim}}\!$\{jan,marcus\} }
\maketitle

\begin{abstract}\noindent
We study the properties of the Minimum Description Length
principle for sequence prediction, considering a two-part MDL
estimator which is chosen from a countable class of models. This
applies in particular to the important case of {\it universal
sequence prediction}, where the model class corresponds to all
algorithms for some fixed universal Turing machine (this
correspondence is by enumerable semimeasures, hence the resulting
models are stochastic). We prove convergence theorems similar to
Solomonoff's theorem of universal induction, which also holds for
general Bayes mixtures. The bound characterizing the convergence
speed for MDL predictions is exponentially larger as compared to
Bayes mixtures. We observe that there are at least {\it three}
different ways of using MDL for prediction. One of these has worse
prediction properties, for which predictions only converge if the
MDL estimator stabilizes. We establish sufficient conditions for
this to occur. Finally, some immediate consequences for complexity
relations and randomness criteria are proven.
\end{abstract}

\begin{keywords}
Minimum Description Length, Sequence Prediction,
Convergence, Discrete Model Classes, Universal Induction,
Stabilization, Algorithmic Information Theory.
\end{keywords}

\newpage
\section{Introduction}

The Minimum Description Length (MDL) principle is one of the most
important concepts in Machine Learning, and serves as a scientific
guide, in general. In particular, the process of building a model
for any kind of given data is governed by the MDL principle in the
majority of cases. The following illustrating example is probably
familiar to many readers: A Bayesian net (or neural network) is
constructed from (trained with) some data. We may just
determine (train) the net in order to fit the data as closely as
possible, then we are describing the data very precisely, but
disregard the description of the net itself. The resulting net is
a maximum likelihood estimator. Alternatively, we may {\it
simultaneously} minimize the ``residual'' description length of the
data given the net {\it and} the description length of the net.
This corresponds to minimizing a {\it regularized} error term, and
the result is a maximum a posteriori or MDL estimator. The latter
way of modelling is not only superior to the former in most
applications, it is also conceptually appealing since it
implements the simplicity principle, Occam's razor.

The MDL method has been studied on all possible levels from very
concrete and highly tuned practical applications up to general
theoretical assertions (see e.g.
\cite{Wallace:68,Rissanen:78,Gruenwald:98}). The aim of this work
is to contribute to the theory of MDL. We regard Bayesian or
neural nets or other models as just some particular class of
models. We identify (probabilistic) models with {\it
(semi)measures}, {\it data} with the initial part of a {\it
sequence} $x_1,x_2,\ldots,x_{t-1}$, and the task of learning with
the problem of {\it predicting} the next symbol
$x_t$ (or more symbols). The sequence $x_1,x_2,\ldots$ itself is
generated by some {\it true} but unknown {\it distribution} $\mu$.

An two-part MDL estimator for some string $x=x_1,\ldots,x_{t-1}$
is then some short description of the semimeasure, while
simultaneously the probability of the data under the related
semimeasure is large.
Surprisingly little work has been done on this general setting of
sequence {\it prediction} with MDL. In contrast, most work
addresses MDL for {\it coding and modeling}, or others, see e.g.
\cite{Barron:98,Rissanen:96,Barron:91,Rissanen:99}. Moreover,
there are some results for the prediction of independently
identically distributed (i.i.d.) sequences, see e.g.
\cite{Barron:91}. There, discrete model classes are considered,
while most of the material available focusses on continuous model
classes. In our work we will study countable classes of {\em
arbitrary} semimeasures.

There is a strong motivation for considering both countable
classes and semimeasures: In order to derive performance
guarantees one has to assume that the model class contains the
true model. So the larger we choose this class, the less
restrictive is this assumption. From a computational point of view
the largest relevant class is the class of {\em all}
lower-semicomputable semimeasures. We call this setup {\it
universal sequence prediction}. This class is at the foundations
of and has been intensely studied in Algorithmic Information
Theory \cite{Zvonkin:70,Li:97,Calude:02}. Since algorithms do not
necessarily halt on each string, one is forced to consider the
more general class of semimeasures, rather than measures.
Solomonoff \cite{Solomonoff:64,Solomonoff:78} defined a universal
induction system, essentially based on a Bayes mixture over this
class (see \cite{Hutter:99errbnd,Hutter:03optisp} for recent
developments). There seems to be no work on MDL for this class,
which this paper intends to change. What has been studied
intensely in \cite{Hutter:03unimdl} is the so called one-part MDL
over the class of deterministic computable models (see also
Section \ref{secCR}).

The paper is structured as follows. Section \ref{secPN}
establishes basic definitions. In Section \ref{secMDL}, we
introduce the MDL estimator and show how it can be used for
sequence prediction in at least three ways. Sections
\ref{secNormDyn} and \ref{secStatic} are devoted to convergence
theorems. In Section \ref{secHybrid}, we study the stabilization
properties of the MDL estimator. The setting of universal sequence
prediction is treated in Section \ref{secCR}. Finally, Section
\ref{secDC} contains the conclusions.

\section{Prerequisites and Notation}\label{secPN}

We build on the notation of \cite{Li:97} and
\cite{Hutter:03unimdl}. Let the alphabet $\XX$ be a finite set of
symbols. We consider the spaces $\XX^*$ and $\XX^\infty$ of finite
strings and infinite sequences over $\XX$. The initial part of a
sequence up to a time $t\in\NNN$ or $t-1\in\NNN$ is denoted by
$x\leqt$ or $x\ltt$, respectively. The empty string is denoted by
$\epstr$.

A {\it semimeasure} is a function $\nu:\XX^*\to[0,1]$ such that
\beq
\label{eq:semimeasure}
\nu(\epstr)\leq 1 \und
\nu(x)\geq \sum_{a\in\XX}\nu(xa) \for_all x\in\XX^*
\eeq
holds. If equality holds in both inequalities of
(\ref{eq:semimeasure}), then we have a {\it measure}. Let $\CC$ be
a countable class of (semi)measures, i.e.\ $\CC=\{\nu_i:i\in I\}$
with finite or infinite index set $I\subseteq\NNN$. A
(semi)measure $\tilde\nu$ {\it dominates} the class $\CC$ iff for
all $\nu_i\in\CC$ there is a constant $c(\nu_i)>0$ such that
$\nu(x)\geq c(\nu_i)\cdot\nu_i(x)$ holds for all $x\in\XX^*$. The
dominant semimeasure $\tilde\nu$ need not be contained in $\CC$,
but if it is, we call it a {\it universal} element of $\CC$.

Let $\CC$ be a countable class of (semi)measures, where each
$\nu\in\CC$ is associated with a weight $w_\nu>0$ and $\sum_{\nu} w_\nu\leq 1$.
We may interpret the weights as a {\it prior} on $\CC$. Then it is
obvious that the Bayes mixture
\beq
\label{eq:xi}
\xi(x)=\xi_{[\CC]}(x)=\sum_{\nu\in\CC}w_\nu \nu(x),\ x\in\XX^*,
\eeq
dominates $\CC$. Assume that there is some measure
$\mu\in\CC$, the {\it true distribution}, generating sequences
$x\ltinf\in\XX^\infty$. Normally $\mu$ is unknown.
(Note that we require $\mu$ to be a measure, while $\CC$ may
contain also semimeasures in general. This is motivated by the
setting of universal sequence prediction as already indicated.) If
some initial part $x\ltt$ of a sequence is given, the probability
of observing $x_t\in\XX$ as a next symbol is given by
\beq
  \label{eq:basicprediction}
  \mu(x_t|x\ltt)=\frac{\mu(x\ltt x_t)}{\mu(x\ltt)} \wenn
  \mu(x\ltt)>0 \ \und\  \mu(x_t|x\ltt)=0 \wenn \mu(x\ltt)=0.
\eeq
The case $\mu(x\ltt)=0$ is stated only for well-definedness, it
has probability zero. Note that $\mu(x_t|x\ltt)$ can depend on
$x\ltt$. We may generally define the quantity (\ref{eq:basicprediction})
for {\it any} function $\ph:\XX^*\to[0,1]$, we call
$\ph(x_t|x\ltt)=\frac{\ph(x\leqt)}{\ph(x\ltt)}$ the {\it
$\ph$-prediction}. Clearly, this is not necessarily a probability
on $\XX$ for general $\ph$. For a semimeasure
$\nu$ in particular, the $\nu$-prediction $\nu(\cdot|x\ltt)$ is a
semimeasure on $\XX$.

We define the {\it expectation} with respect to the true
probability $\mu$: Let $n\geq 0$ and $f:\XX^n\to\RRR$ be a
function, then
\beq
\label{eq:expectation}
\E\ f=\E\ f(x\leqn)=\sum_{x\leqn\in\XX^n}\mu(x\leqn)f(x\leqn).
\eeq
Generally, we may also define the expectation as an integral over
infinite sequences. But since we won't need it, we can keep things
simple. We can now state a central result about prediction with
Bayes mixtures in a form independent of Algorithmic Information
Theory.

\begin{Theorem}
\label{Theorem:Solomonoff}
For any class of (semi)measures $\CC$ containing the true
distribution $\mu$ and any $n\geq 1$, we have
\beq
\label{eq:Solomonoff}
\sum_{t=1}^n \E \sum_{a\in\XX}
  \Big( \mu(a|x\ltt)-\xi(a|x\ltt) \Big)^2 \ \leq \ \ln w_\mu^{-1}.
\eeq
\end{Theorem}

This was found by Solomonoff (\cite{Solomonoff:78}) for universal
sequence prediction. A proof is also given in \cite{Li:97} (only
for binary alphabet) or \cite{Hutter:01alpha} (arbitrary
alphabet). It is surprisingly simple once Lemma
\ref{Lemma:EntropyIneq} is known. A few lines analogous to
(\ref{Eq:TheoremRhoConvergesIMS0}) and
(\ref{Eq:TheoremRhoConvergesIMS00}) exploiting the dominance of
$\xi$ are sufficient.

The bound (\ref{eq:Solomonoff}) asserts convergence of the
$\xi$-predictions to the $\mu$-predictions {\it in mean sum
(i.m.s.)}, since we define
\beq
\label{eq:IMS}
\ph\toims\mu\quad\Longleftrightarrow\quad
\exists\ C>0:\ \sum_{t=1}^\infty \E \sum_{a\in\XX}
  \Big( \mu(a|x\ltt)-\ph(a|x\ltt) \Big)^2 \leq C.
\eeq
Convergence i.m.s.\ implies convergence with
$\mu$-probability one (w.$\mu$-p.1), since otherwise the
sum would be infinite. Moreover, convergence i.m.s.\ provides a
rate or speed of convergence in the sense that the expected number
of times $t$ in which $\ph(a|x\ltt)$ deviates more than
$\eps$ from $\mu(a|x\ltt)$ is finite and bounded by $C/\eps^2$ and the
probability that the number of $\eps$-deviations exceeds
$C\over\eps^2\delta$ is smaller than $\delta$. If the
quadratic differences were monotonically decreasing (which is
usually not the case), we could even conclude convergence faster
than $\frac{1}{t}$.

\begin{MyPar}{Probabilities vs.\ Description Lengths}
\label{par:prob}
\begin{sloppy}
By the Kraft inequality, each (semi)measure can be associated with
a code length or {\it complexity} by means of the negative
logarithm, where all (binary) codewords form a prefix-free set.
The converse holds as well. E.g. for the weights $w_\nu$ with
$\sum w_\nu\leq 1$, codes of lengths $\lceil-\lb w_\nu\rceil$ can
be found. It is often only a matter of notational convenience if
description lengths or probabilities are used, but description
lengths are generally preferred in Algorithmic Information Theory.
Keeping the equivalence in mind, we will develop the general
theory in terms of probabilities, but formulate parts of the
results in universal sequence prediction rather in terms of
complexities.
\end{sloppy}
\end{MyPar}

\section{MDL Estimator and Predictions}\label{secMDL}

Assume that $\CC$ is a countable class of semimeasures together
with weights $(w_\nu)_{\nu\in\CC}$, and $x\in\XX^*$ is some
string. Then the {\it maximizing element} $\nu^x$, often called
MAP estimator, is defined as
\beqn
  \nu^x = \nu^x_{[\CC]}=\arg\max_{\nu\in\CC}\{w_\nu\nu(x)\}.
\eeqn
In fact the maximum is attained since for each $\eps\in(0,1)$ only
a finite number of elements fulfil $w_\nu\nu(x)>\eps$. Observe
immediately the correspondence in terms of {\it description
lengths} rather than {\it probabilities}:
$\nu^x = \arg\min_{\nu\in\CC}\big\{-\lb w(\nu)-\lb \nu(x)\big\}$.
Then the {\it minimum description length principle} is obvious:
$\nu^x$ minimizes the joint description length of the model plus
the data given the model%
\footnote{Precisely, we define a MAP (maximum a posteriori)
estimator. For two reasons, information theorists and
statisticians would not consider our definition as MDL in the
strong sense. First, MDL is often associated with a specific
prior. Second, when coding some data $x$, one can exploit the fact
that once the model $\nu^x$ is specified, only data which leads to
the maximizing element $\nu^x$ needs to be considered. This allows
for a description shorter than $\lb \nu^x(x)$. Since however most
authors refer to MDL, we will keep using this general term instead
of MAP, too.} (see the last paragraph of the previous section). As
explained before, we stick to the product notation.

For notational simplicity we set $\nu^*(x)=\nu^x(x)$. The {\it
two-part MDL estimator} is defined by
\beqn
  \rrho(x) = \rrho_{[\CC]}(x) = w_{\nu^x}\nu^x(x) = \max_{\nu\in\CC}\{w_\nu \nu(x)\}.
\eeqn
So $\rrho$ chooses the maximizing element with respect to its
argument. We may also use the version $\rrho^y(x) :=
w_{\nu^y}\nu^y(x)$ for which the choice depends on the superscript
instead of the argument. For each $x,y\in\XX^*$,
$\xi(x)\geq\rrho(x)\geq\rrho^y(x)$
is immediate.

We can define MDL predictors according to
(\ref{eq:basicprediction}). There are {\it at least three}
possible ways to use MDL for prediction.

\begin{Def}
\label{Def:DynamicMDL}
The {\em dynamic} MDL predictor is defined as
\beqn
  \rrho(a|x) = \frac{\rrho(xa)}{\rrho(x)}
  = \frac{\rrho^{xa}(xa)}{\rrho^x(x)}.
\eeqn
That is, we look for a short description of $xa$ and relate it to
a short description of $x=x\ltt$. We call this dynamic since for
each possible $a$ we have to find a new MDL estimator. This is the
closest correspondence to the $\xi$-predictor.
\end{Def}

\begin{Def}
\label{Def:StaticMDL}
The {\em static} MDL predictor is given by
\beqn
  \rrho^{\mathrm{\mathrm{static}}}(a|x)
  = \rrho^x(a|x) = \frac{\rrho^x(xa)}{\rrho(x)}
  = \frac{\rrho^x(xa)}{\rrho^x(x)}
  = \frac{\nu^x(xa)}{\nu^x(x)}.
\eeqn
Here obviously only {\it one} MDL estimator $\rrho^x$ has to be
identified, which may be more efficient in practice.
\end{Def}

\begin{Def}
\label{Def:HybridMDL}
The {\em hybrid} MDL predictor is given by
$
  \rrho^{\mathrm{hyb}}(a|x)
  = \frac{\nu^*(xa)}{\nu^*(x)}
$.
This can be paraphrased as ``do dynamic MDL and drop the weights".
It is somewhat in-between static and dynamic MDL.
\end{Def}

The range of the static MDL predictor is obviously contained in
$[0,1]$. For the dynamic MDL predictor, this holds by
$\rrho^x(x)\geq\rrho^{xa}(x)\geq\rrho^{xa}(xa)$,
while for the hybrid MDL predictor it is generally false.

Static MDL is omnipresent in machine learning and applications. In
fact, many common prediction algorithms can be abstractly
understood as static MDL, or rather as approximations. Namely, if
a prediction task is accomplished by building a {\it model} such
as a neural network with a suitable regularization to prevent
``overfitting", this is just searching an MDL estimator within a
certain class of distributions. After that, only this model is
used for prediction. Dynamic and hybrid MDL are applied more
rarely due to their larger computational effort. For example, the
similarity metric proposed in \cite{Li:03} can be interpreted as
(a deterministic variant of) dynamic MDL. For hybrid MDL, we will
see that the prediction properties are worse than for dynamic and
static MDL.

We will need to convert our MDL predictors to {\it measures} on
$\XX$ by means of {\it normalization}. If $\ph:\XX^*\to[0,1]$ is any
function, then
\beqn
  \ph\_norm(a|x\ltt)
  \ = \ \frac{\ph(a|x\ltt)}{\sum_{a'\in\XX}\ph(a'|x\ltt)}
  \ = \ \frac{\ph(x\ltt a)}{\sum_{a'\in\XX}\ph(x\ltt a')}
\eeqn
(assume that the denominator is different from zero, which is
always true with probability 1 if $\ph$ is an MDL predictor). This
procedure is known as {\it Solomonoff normalization}
(\cite{Solomonoff:78,Li:97}) and results in
$\nu\_norm(x\leqn) = \nu(x\leqn)/\big[\nu(\epstr)N_\nu(x_{<n})\big]$,
where
\beq\label{Eq:Normalizer}
  N_\nu(x) = \prod_{t=1}^{\l(x)+1} \frac{\sum_{a\in\XX}\nu(x\ltt a)}{\nu(x\ltt)}
\eeq
is the normalizer. Before proceeding with the theory, an example
is in order.
\begin{Example}
\label{ex:iid}
Let $n\in\NNN$, $\XX=\{1,\ldots,n\}$, and
\beqn
  \CC=\Big\{\nu_\th(x_{1:t})=\th_{x_1}\!\cdot...\!\cdot\th_{x_t} : \th\in\Theta\Big\}
  \qmbox{with} \Theta=\Big\{\th\in([0,1]\cap\QQQ)^n:\sum_{i=1}^n \th_i=1\Big\}
\eeqn
be the set of all rational probability vectors with any prior
$(w_\th)_{\th\in\Theta}$. Each $\th\in\Theta$ generates sequences $x\ltinf$
of {\it independently identically distributed (i.i.d)} random
variables such that $P(x_t=i)=\th_i$ for all $t\geq 1$ and
$1\leq i\leq n$. If $x\leqt$ is the initial part of a sequence and
$\alpha\in\Theta$ is defined by $\alpha_i=|\{s\leq t:x_s=i\}|$,
then it is easy to see that
\beqn
\nu^{x\leqt}=\arg\max_{\th\in\Theta}\left\{ w(\th)\cdot\exp\big[
-t\!\cdot\!D(\alpha\|\th)\big]\right\},
\eeqn
where $D(\alpha\|\th)= \sum_{i=1}^n
\alpha_i\ln\frac{\alpha_i}{\th_i}$ is the {\it Kullback-Leibler
divergence}. If $|\XX|=2$, then $\Theta$ is also called a {\it
Bernoulli class}, and one usually takes the binary alphabet
$\XX=\BBB=\{0,1\}$ in this case.
\end{Example}

\section{Dynamic MDL}\label{secNormDyn}

We can start to develop results. It is surprisingly easy to give a
convergence proof w.p.1 of the non-normalized dynamic MDL
predictions based on martingales. However we omit it, since it
does not include a convergence speed assertion as i.m.s.\ results
do, nor does it yield an off-sequence statement about
$\rrho(a|x\ltt)$ for $a\neq x_t$ which is necessary for
prediction.

\begin{Lemma} \label{Lemma:DiffIsSemimeasure}
For an arbitrary class of (semi)measures $\CC$, we have
\bqan
& (i) &\  \rrho(x)-\sum_{a\in\XX}\rrho(xa)\ \leq\
  \xi(x)-\sum_{a\in\XX}\xi(xa) {\rm\ and}\\
& (ii) &\  \rrho^x(x)-\sum_{a\in\XX}\rrho^x(xa)\ \leq\
  \xi(x)-\sum_{a\in\XX}\xi(xa)
\eqan
for all $x\in\XX^*$. In particular,
$\xi-\rrho$ is a semimeasure.
\end{Lemma}

\begin{Proof}
For all $x\in\XX^*$, with $f:=\xi-\rrho$ we have
\bqan
  \sum_{a\in\XX} f(xa)
  & = & \sum_{a\in\XX} \Big(\xi(xa)-\rrho(xa)\Big)
  \leq \sum_{a\in\XX} \Big(\xi(xa)-\rrho^x(xa)\Big)\\
  & =& \sum_{\nu\in\MM\setminus\{\nu^x\}} \sum_{a\in\XX} w_\nu\nu(xa)
  \leq \sum_{\nu\in\MM\setminus\{\nu^x\}} w_\nu\nu(x)
  = \xi(x)-\rrho(x)
  = f(x).
\eqan
The first inequality follows from $\rrho^x(xa)\leq \rrho(xa)$, and
the second one holds since all $\nu$ are semimeasures. Finally,
$f(x)=\xi(x)-\rrho(x)=\sum_{\nu\in\MM\setminus\{\nu^x\}}
w_\nu\nu(x) \geq 0$ and
$f(\epstr)=\xi(\epstr)-\rrho(\epstr)\leq 1$. Hence $f$ is a
semimeasure.
\end{Proof}

\begin{Lemma} \label{Lemma:EntropyIneq}
Let $\mu$ and $\tilde\mu$ be measures on $\XX$, then
\beqn
  \sum_{a\in\XX}\big(\mu(a)-\tilde\mu(a)\big)^2 \leq
  \sum_{a\in\XX} \mu(a)\ln \frac{\mu(a)}{\tilde\mu(a)}.
\eeqn
\end{Lemma}

See e.g.\ \cite[Sec.3.2]{Hutter:01alpha} for a proof.

\begin{Theorem} \label{Theorem:RhoConvergesIMS}
For any class of (semi)measures $\CC$ containing the true
distribution $\mu$ and for all $n\in\NNN$, we have
\beqn
  \sum_{t=1}^n \ \E \sum_{a\in\XX} \big(\mu(a|x\ltt)-\rrho\_norm(a|x\ltt)\big)^2
  \leq w_\mu^{-1} + \ln w_\mu^{-1}.
\eeqn
That is, $\rrho\_norm(a|x\ltt)\toims\mu(a|x\ltt)$ (see
(\ref{eq:IMS})), which implies
$\rrho\_norm(a|x\ltt)\to\mu(a|x\ltt)$ with $\mu$-probability one.
\end{Theorem}

\begin{Proof}
From Lemma \ref{Lemma:EntropyIneq}, we know
\bqa
  \nonumber
  \lefteqn{\sum_{t=1}^n \ \E \sum_{a\in\XX} \big(\mu(a|x\ltt)-\rrho\_norm(a|x\ltt)\big)^2
  \leq \sum_{t=1}^n \ \E \sum_{a\in\XX} \mu(a|x\ltt)
  \ln \frac{\mu(a|x\ltt)}{\rrho\_norm(a|x\ltt)}
  }\\
  & = & \sum_{t=1}^n \ \E \ln \frac{\mu(x_t|x\ltt)}{\rrho\_norm(x_t|x\ltt)}
  = \sum_{t=1}^n \ \E \left[
  \ln\frac{\mu(x_t|x\ltt)}{\rrho(x_t|x\ltt)}
  + \ln \frac{\sum_{a\in\XX}\rrho(x\ltt
  a)}{\rrho(x\ltt)}\right].\quad
  \label{Eq:TheoremRhoConvergesIMS0}
\eqa
Then we can estimate
\beq \label{Eq:TheoremRhoConvergesIMS00}
  \sum_{t=1}^n \ \E \ln\frac{\mu(x_t|x\ltt)}{\rrho(x_t|x\ltt)}
  \ = \ \E\ \ln\prod_{t=1}^n \frac{\mu(x_t|x\ltt)}{\rrho(x_t|x\ltt)}
  \ = \ \E\ \ln\frac{\mu(x\leqn)}{\rrho(x\leqn)}
  \ \leq \ \ln w_\mu^{-1},
\eeq
since always $\frac{\mu}{\rrho}\leq w_\mu^{-1}$. Moreover, by
setting $x=x\ltt$, using $\ln u\leq u-1$, adding an always
positive max-term, and finally using
$\frac{\mu}{\rrho}\leq w_\mu^{-1}$ again, we obtain
\bqa
\nonumber
\lefteqn{
  \E \ \ln \frac{\sum_a\rrho(x\ltt a)}{\rrho(x\ltt)}
  \leq \E\left[\frac{\sum_a\rrho(xa)}{\rrho(x)}-1\right]
   = \sum_{\zwidths{\l(x)=t-1}}
   \frac{\mu(x)\Big[\big(\sum_a\rrho(xa)\big)-\rrho(x)\Big]}{\rrho(x)}
} \\
\nonumber
  & \leq &
  \sum_{\zwidths{\l(x)=t-1}}
\frac{\mu(x)\Big[\left(\sum_{a\in\XX}\rrho(xa)\right)-\rrho(x)+
  \max\left\{0,\rrho(x)-\sum_{a\in\XX}\rrho(xa)\right\}\Big]}{\rrho(x)}
\\
\label{Eq:TheoremRhoConvergesIMS1}
& \leq &
  w_\mu^{-1}
  \sum_{\l(x)=t-1} \left[\Big(\sum_{a\in\XX}\rrho(xa)\Big)-\rrho(x)+
  \max\Big\{0,\rrho(x)-\sum_{a\in\XX}\rrho(xa)\Big\}\right] .
\eqa
We proceed by observing
\beq \label{Eq:TheoremRhoConvergesIMS2}
  \sum_{t=1}^n \sum_{\l(x)=t-1}\Big[\Big(\sum_{a\in\XX}\rrho(xa)\Big)-\rrho(x)\Big]
  = \sum_{t=1}^n\Big[\sum_{\zwidths{\l(x)=t}}\rrho(x)-
  \sum_{\zwidths{\l(x)=t-1}}\rrho(x)\Big]
  = \Big[\sum_{\zwidths{\l(x)=n}}\rrho(x)\Big]-\rrho(\epstr)
\eeq
which is true since for successive $t$ the positive and negative
terms cancel. From Lemma \ref{Lemma:DiffIsSemimeasure} we know
$  \rrho(x)-\sum_{a\in\XX}\rrho(xa)\ \leq\
  \xi(x)-\sum_{a\in\XX}\xi(xa)$
and therefore
\bqa
  \nonumber
  \sum_{t=1}^n \sum_{\zwidths{\quad \l(x)=t-1}}
   \max\Big\{0,\rrho(x)-\sum_{a\in\XX}\rrho(xa)\Big\}
  & \leq & \sum_{t=1}^n \sum_{\zwidths{\quad \l(x)=t-1}}
  \max\Big\{0,\xi(x)-\sum_{a\in\XX}\xi(xa)\Big\}
  \\
  = \sum_{t=1}^n \sum_{\zwidths{\quad \l(x)=t-1}}\Big[\xi(x)-\sum_{a\in\XX}\xi(xa)\Big]
  & = &  \xi(\epstr)-\sum_{\l(x)=n}\xi(x).
  \label{Eq:TheoremRhoConvergesIMS3}
\eqa
Here we have again used the fact that positive and negative terms
cancel for successive $t$, and moreover the fact that $\xi$ is a
semimeasure. Combining (\ref{Eq:TheoremRhoConvergesIMS1}),
(\ref{Eq:TheoremRhoConvergesIMS2}) and
(\ref{Eq:TheoremRhoConvergesIMS3}), and observing
$\rrho\leq\xi\leq 1$, we obtain
\beq \label{Eq:TheoremRhoConvergesIMSNormalizer}
  \sum_{t=1}^n \ \E \ln \frac{\sum_a\rrho(x\ltt a)}{\rrho(x\ltt)}
\leq w_\mu^{-1}
   \left[\xi(\epstr)-\rrho(\epstr)+\sum_{\zwidths{\l(x)=n}}\big(\rrho(x)-\xi(x)\big)\right]
  \leq w_\mu^{-1}\xi(\epstr) \leq w_\mu^{-1}.
\eeq
Therefore, (\ref{Eq:TheoremRhoConvergesIMS0}), (\ref{Eq:TheoremRhoConvergesIMS00})
and (\ref{Eq:TheoremRhoConvergesIMSNormalizer}) finally prove the assertion.
\end{Proof}

This is the first convergence result in mean sum, see
(\ref{eq:IMS}). It implies both on-sequence and off-sequence
convergence. Moreover, it asserts the convergence is ``fast" in
the sense that the sum of the total expected deviations is bounded
by $w_\mu^{-1}+\ln w_\mu^{-1}$. Of course, $w_\mu^{-1}$ can be
very large, namely $2$ to the power of complexity of $\mu$. The
following example will show that this bound is sharp (save for a
constant factor). Observe that in the corresponding result for
mixtures, Theorem \ref{Theorem:Solomonoff}, the bound is much
smaller, namely $\ln w_\mu^{-1}\ =$ complexity of $\mu$.

\begin{Example}\label{Ex:LowerBound}
Let $\XX=\{0,1\}$, $N\geq 1$ and
$\CC=\{\nu_1,\ldots,\nu_{N-1},\mu\}$. Each $\nu_i$ is a
deterministic measure concentrated on the sequence
$1^{i-1}0^\infty$, while the true distribution $\mu$
is deterministic and concentrated on $x\ltinf=1^\infty$. Let
$w_{\nu_i}=w_\mu=\frac{1}{N}$ for all $i$. Then $\mu$ generates
$x\ltinf$, and for each $t\leq N-1$ we have
$\rrho\_norm(0|x\ltt)=\rrho\_norm(1|x\ltt)=\frac{1}{2}$. Hence,
$\sum_t\E\sum_a\big(\mu(a|x\ltt)-\rrho\_norm(a|x\ltt\big))^2=\frac{1}{2}(N-1)\approx
\frac{1}{2}w_\mu^{-1}$ for large $N$.
Here, $\mu$ is Bernoulli, while the $\nu_i$ are not. It might be
surprising at a first glance that there are even classes $\CC$
containing \emph{only} Bernoulli distributions, where the
exponential bound is sharp \cite{Poland:04mdlspeed}.
\end{Example}

\begin{Theorem} \label{thNormalizer} \label{thRhoNorm}
For any class of (semi)measures $\CC$ containing the true
distribution $\mu$, we have
\bqan
  & (i) &
  \sum_{t=1}^\infty \E \left|\,
  \ln \sum_{a\in\XX}\rrho(a|x\ltt) \right|
  \ \leq \ 2 w_\mu^{-1} {\rm\quad and}
\\
  & (ii) & \sum_{t=1}^\infty \E \sum_{a\in\XX}
  \Big| \rrho\_norm(a|x\ltt)-\rrho(a|x\ltt) \Big|
  \ = \
  \sum_{t=1}^\infty \E
  \Big| 1- \sum_{a\in\XX}\rrho(a|x\ltt) \Big|
  \ \leq \ 2 w_\mu^{-1}.
\eqan
Consequently, $\rrho(a|x\ltt)\toims\mu(a|x\ltt)$, and for almost
all $x_{<\infty}\in\XX^\infty$, the normalizer $N_\rrho$ defined
in (\ref{Eq:Normalizer}) converges to a number which is finite and
greater than zero, i.e.\ $0<N_\rrho(x_{<\infty})<\infty$.
\end{Theorem}

\begin{Proof}
$(i)$ Define $u^+=\max\{0,u\}$ for $u\in\RRR$, then for $x:=x\ltt\in\XX^{t-1}$ we have
\bqan
  \E \Big|\ln \sum_{a\in\XX}&\!\rrho(a|x)&\!\! \Big| =
  \E \left|
  \ln {\sum_a\rrho(xa)\over\rrho(x)} \right|
  = \E \left[ \Big(\ln{\sum_a\rrho(xa)\over\rrho(x)}\Big)^+
  +\Big(\ln{\rrho(x)\over\sum_a\rrho(xa)}\Big)^+\right]
\\
  & \leq & \E {\big(\sum_a\rrho(xa)-\rrho(x)\big)^+\over\rrho(x)}
  \ + \ \E {\big(\rrho(x)-\sum_a\rrho(xa)\big)^+\over\sum_a\rrho(xa)}
\\
  & = & \sum_{\zwidths{\l(x)=t-1}} {\mu(x)\big(\sum_a\rrho(xa)-\rrho(x)\big)^+\over\rrho(x)}
  \ + \sum_{\zwidths{\l(x)=t-1}} {\mu(x)\big(\rrho(x)-\sum_a\rrho(xa)\big)^+\over\sum_a\rrho(xa)}
\\
  & \leq & w_\mu^{-1} \sum_{\zwidths{\l(x)=t-1}}
  \big({\textstyle\sum_a\rrho(xa)}-\rrho(x)\big)^+
  \ + \ w_\mu^{-1} \sum_{\zwidths{\l(x)=t-1}}
  \big(\rrho(x)-{\textstyle\sum_a\rrho(xa)}\big)^+
\\
  & = & w_\mu^{-1} \sum_{\zwidths{\l(x)=t-1}} \left[\textstyle\sum_a\rrho(xa)-\rrho(x)
  +
  2\big(\rrho(x)-{\textstyle\sum_a\rrho(xa)}\big)^+\right].
\eqan
Here, $|u|=u^++(-u)^+=-u+2u^+$, $\ln u\leq u-1$, and
$\rrho\geq w_\mu\mu$ have been used, the latter implies also
$\sum_a\rrho(xa)\geq w_\mu\sum_a\mu(xa)=w_\mu\mu(x)$.
The last expression in this (in)equality chain, when summed over
$t=1...\infty$ is bounded by $2w_\mu^{-1}$ by essentially the same
arguments (\ref{Eq:TheoremRhoConvergesIMS1}) -
(\ref{Eq:TheoremRhoConvergesIMSNormalizer}) as in the proof of
Theorem \ref{Theorem:RhoConvergesIMS}.

$(ii)$ Let again $x:=x\ltt$ and use
$\rrho\_norm(a|x)=\rrho(a|x)/\sum_b\rrho(b|x)$ to obtain
\bqan
  \sum_a \Big| \rrho\_norm(a|x)-\rrho(a|x) \Big|
  & = & \sum_a {\rrho(a|x)\over\sum_b\rrho(b|x)}\Big| 1-\sum_b\rrho(b|x) \Big|
   =  \Big| 1-\sum_b\rrho(b|x) \Big|\\
  & = & {\big(\sum_a\rrho(xa)-\rrho(x)\big)^+\over\rrho(x)}
   +  {\big(\rrho(x)-\sum_a\rrho(xa)\big)^+\over\rrho(x)}
\eqan
Then take the expectation $\E$ and the sum
$\sum_{t=1}^\infty$ and proceed as in $(i)$.
Finally, $\rrho(a|x\ltt)\toims\mu(a|x\ltt)$ follows by combining
$(ii)$ with Theorem \ref{Theorem:RhoConvergesIMS}, and by $(i)$, $
\sum_1^n \big| \ln \frac{\sum_a\rrho(x\ltt a)}{\rrho(x\ltt)} \big|
$ is bounded in $n$ with $\mu$-probability 1, thus the same is
true for $\ln N_\rrho(x_{<\infty}) = \sum_1^\infty \ln
\frac{\sum_{a\in\XX}\rrho(x\ltt a)}{\rrho(x\ltt)}$.
\end{Proof}

\section{Static MDL}\label{secStatic}

So far, we have considered dynamic MDL from Definition
\ref{Def:DynamicMDL}. We turn now to the static variant
(Definition \ref{Def:StaticMDL}), which is usually more efficient and thus
preferred in practice.

\begin{Theorem} \label{thSMDLBound}
For any class of (semi)measures $\CC$ containing the true
distribution $\mu$, we have
\beqn
  \sum_{t=1}^\infty \E
  \sum_{a\in\XX} \Big| \rrho\_norm^{x\ltt}(a|x\ltt) - \rrho^{x\ltt}(a|x\ltt) \Big|
  \ = \ \sum_{t=1}^\infty \E
  \Big| 1- \sum_{a\in\XX}\rrho^{x\ltt}(a|x\ltt) \Big|
  \ \leq \ w_\mu^{-1}.
\eeqn
\end{Theorem}

\begin{Proof}
We proceed in a similar way as in the proof of Theorem
\ref{Theorem:RhoConvergesIMS},
(\ref{Eq:TheoremRhoConvergesIMS1}) - (\ref{Eq:TheoremRhoConvergesIMS3}).
From Lemma \ref{Lemma:DiffIsSemimeasure}, we know
$  \rrho(x)-\sum_a\rrho^x(xa)\ \leq\  \xi(x)-\sum_a\xi(xa)$.
Then
\bqan
  \lefteqn{
  \sum_{t=1}^n \E
  \Big| 1- \sum_{a\in\XX}\rrho^{x\ltt}(a|x\ltt) \Big|
  = \sum_{t=1}^n\ \E \frac{\rrho(x\ltt)-\sum_{a\in\XX}\rrho^{x\ltt}(x\ltt
  a)}{\rrho(x\ltt)}
  }
\\
  & = & \sum_{t=1}^n \sum_{\l(x)=t-1} \!\!\!\!\mu(x)
  \frac{\rrho(x)-\sum_{a\in\XX}\rrho^x(xa)}{\rrho(x)}
  \leq w_\mu^{-1}\ \sum_{t=1}^n \sum_{\l(x)=t-1}
  \!\!\!\!\!\!\Big[\rrho(x)-\sum_{a\in\XX}\rrho^x(xa)\Big]
\\
  & \leq & w_\mu^{-1}\ \sum_{t=1}^n \sum_{\l(x)=t-1}
  \left[\xi(x)-\sum_{a\in\XX}\xi(xa)\right]
  \leq w_\mu^{-1}\left[\xi(\epstr)-\sum_{\l(x)=n}\xi(x)\right]
  \ \leq \ w_\mu^{-1}.
\eqan
for all $n\in\NNN$. This implies the assertion. Again we have used
$\frac{\mu}{\rrho}\leq w_\mu^{-1}$ and the fact that positive and
negative terms cancel for successive $t$.
\end{Proof}

\begin{Cor}
\label{Cor:MuRhoAll}
Let $\CC$ contain the true distribution $\mu$, then%
\beqn
\begin{array}{r l @{\quad} c @{\quad} l}
  \sum_t \E \sum_a & \big( \mu(a|x\ltt)-\rrho\_norm(a|x\ltt) \big)^2 & \leq & 2 w_\mu^{-1}, \\[3pt]
  \sum_t \E \sum_a & \big( \mu(a|x\ltt)-\rrho(a|x\ltt) \big)^2 & \leq & 8 w_\mu^{-1}, \\[3pt]
  \sum_t \E \sum_a & \big( \mu(a|x\ltt)-\rrho^{x\ltt}(a|x\ltt) \big)^2 & \leq & 21 w_\mu^{-1},\\[3pt]
  \sum_t \E \sum_a & \big( \mu(a|x\ltt)-\rrho^{x\ltt}\_norm(a|x\ltt) \big)^2 & \leq & 32 w_\mu^{-1}.
\end{array}
\eeqn
\end{Cor}
\begin{Proof}
This follows by combining the assertions of Theorems
\ref{Theorem:RhoConvergesIMS} - \ref{thSMDLBound} with the
triangle inequality. For static MDL, use in addition
$ \sum_a\big| \rrho(a|x)-\rrho^x(a|x) \big|
 = \big| \sum_a\rrho(a|x)- \sum_a\rrho^x(a|x) \big|
 \leq \big| \sum_a\rrho(a|x)- 1 \big|
      + \big| 1 - \sum_a\rrho^x(a|x) \big|$
which follows from $\rrho(xa)\geq\rrho^x(xa)$.
\end{Proof}

This corollary recapitulates our results and states convergence
i.m.s (and therefore also with $\mu$-probability 1) for all
combinations of
un-normalized/nor\-malized and dynamic/static MDL predictions.%
\footnote{
We briefly discuss the choice of the total expected square error
for measuring speed of convergence. The expected Kullback-Leibler
distance may seem more natural in the light of our proofs.
However, this quantity behaves well only under dynamic MDL, not
static MDL. To see this, let $\CC$ be the class of all computable
Bernoulli distributions and $\mu$ the measure having
$\mu(0)=\mu(1)=\frac{1}{2}$. Then the sequence
$x=0^n$ has nonzero probability. For sufficiently large $n$,
$\nu^x=\nu_0$ holds (typically already for small $n$), where
$\nu_0$ is the distribution generating only $0$. Then
$D(\mu\|\nu^x)=\infty$, and the expectation is $\infty$, too. The
quadratic distance behaves locally like the Kullback-Leibler
distance (Lemma \ref{Lemma:EntropyIneq}), but otherwise is bounded
and thus more convenient.}

\section{Hybrid MDL and Stabilization}\label{secHybrid}

We now turn to the hybrid MDL variant (see Definition
\ref{Def:HybridMDL}). So far we have not cared about what happens
if two or more (semi)measures obtain the same value $w_\nu
\nu(x)$ for some string $x$. In fact, for the previous results,
the {\it tie-breaking strategy} can be completely arbitrary. This
need not be so for all thinkable prediction methods other than
static and dynamic MDL, as the following example shows.

\begin{Example}
Let $\XX=\BBB$ and $\CC$ contain only two measures, the uniform
measure $\lambda$ which is defined by $\lambda(x)=2^{-\l(x)}$, and
another measure
$\nu$ having $\nu(1x)=2^{-\l(x)}$ and $\nu(0x)=0$. The respective
weights are $w_\lambda=\frac{2}{3}$ and $w_\nu=\frac{1}{3}$. Then,
for each $x$ starting with $1$, we have
$w_\nu\nu(x)=w_\lambda\lambda(x)=\frac{1}{3}2^{-\l(x)+1}$. Therefore,
for all $x\ltinf$ starting with $1$ (a set which has uniform
measure $\frac{1}{2}$), we have a tie. If the maximizing element
$\nu^*$ is chosen to be $\lambda$ for even $t$ and $\nu$ for odd
$t$, then both static and dynamic MDL constantly predict probabilities
of $\frac{1}{2}$ for all $a\in\BBB$. However, the hybrid MDL
predictor values
$\frac{\nu^*(x\ltt a)}{\nu^*(x\ltt)}$ oscillate between ${1\over
4}$ and $1$.
\end{Example}

If the ambiguity in the tie-breaking process is removed, e.g.\ if
always the measure with the larger weight $w_\nu$ is been chosen,
then the hybrid MDL predictor {\it does} converge for this
example. If there are more (semi)measures in the class and there
remains still a tie of shortest programs, an arbitrary program can
be selected, since then the respective measures are equal, too. In
the following, we assume that this tie-breaking rule is applied.

Do the hybrid MDL predictions always converge then? This is
equivalent to asking if the process of selecting a maximizing
element eventually {\it stabilizes}. If there is no stabilization,
then hybrid MDL will necessarily fail as soon as the weights are
not equal. A possible counterexample could consist of two measures
the fraction of which oscillates perpetually around a certain
value. This can indeed happen.

\begin{Example} \label{Example:NotStabilizeApprox}
Let $\XX$ be binary, $\mu(x)=\prod_{i=1}^{\l(x)}\mu_i(x_i)$ and
$\nu(x)=\prod_{i=1}^{\l(x)}\nu_i(x_i)$ with
\beqn
\mu_i(1)=1-2^{-2\left\lceil\frac{i}{2}\right\rceil} {\rm\ and\ }
\nu_i(1)=1-2^{-2\left\lceil\frac{i+1}{2}\right\rceil+1}.
\eeqn
Then one can easily see that
$\mu(111\ldots)=\prod_1^\infty\mu_i(1)>0$,
$\nu(111\ldots)=\prod_1^\infty\nu_i(1)>0$,
and $\frac{\nu(111\ldots)}{\mu(111\ldots)}$ is convergent but
oscillates around its limit. Therefore, we can set $w_\mu$ and
$w_\nu$ appropriately to prevent the maximizing element from
stabilizing on $x\ltinf=111\ldots$ (Moreover, each sequence having
positive measure under $\mu$ and $\nu$ contains eventually only
ones, and the quotient oscillates.)
\end{Example}

The reason for the oscillation in this example is the fact that
measures $\mu$ and $\nu$ are asymptotically very similar. One can
also achieve a similar effect by constructing a measure which is
\emph{dependent} on the past. This shows in particular that we need
both parts of the following definition which states properties
sufficient for a positive result.

\begin{Def} \label{Def:FactorAndUniform}
$(i)$ A (semi)measure $\nu$ on $\XX^\infty$ is called {\em factorizable}
if there are (semi)measures $\nu_i$ on $\XX$ such that
$\nu(x)=\prod_{i=1}^{\l(x)}\nu_i(x_i)$
for all $x\in\XX^*$. That is, the symbols of sequences $x\ltinf$
generated by $\nu$ are independent.\\
$(ii)$ A factorizable (semi)measure
$\mu=\prod\mu_i$ is called {\em uniformly stochastic}, if there is
some $\delta>0$ such that at each time $i$ the probability of all
symbols $a\in\XX$ is either 0 or at least $\delta$. That is,
$\mu_i(a)>0 \Rightarrow \mu_i(a)\geq \delta$ for all $a\in\XX$ and $i\geq 1$.
\end{Def}

In particular, all deterministic measures are uniformly
stochastic. Another simple example of a uniformly stochastic
measure is a probability distribution which generates alternately
random bits by fair coin flips and the digits of the binary
representation of $\pi$.

\begin{Theorem} \label{Theorem:MDLstabilizes}
Let $\CC$ be a countable class of factorizable (semi)measures and
$\mu$ be uniformly stochastic. Then the maximizing element stabilizes
almost surely.
\end{Theorem}

We omit the proof. So in particular, under the conditions of
Theorem \ref{Theorem:MDLstabilizes}, the hybrid MDL predictions
converge almost surely. No statement about the convergence speed
can be made.

\section{Complexities and Randomness}\label{secCR}

In this section, we concentrate on universal sequence prediction.
It was mentioned already in the introduction that this is one
interesting application of the theory developed so far. So
$\CC=\MM$ is the countable set of all enumerable (i.e.\ lower
semicomputable) semimeasures on $\XX^*$. (Algorithms are
identified with semimeasures rather than measures since they need
not terminate.)
$\MM$ contains stochastic models in general, and in particular all
models for computable deterministic sequences. One can show that
this class $\MM$ is determined by {\it all} algorithms on some
fixed universal monotone Turing machine $U$ \cite[Th.
4.5.2]{Li:97}. By this correspondence, each semimeasure
$\nu\in\MM$ is assigned a \emph{canonical weight} $w_\nu=2^{-K(\nu)}$
(where $K(\nu)$ is the Kolmogorov complexity of $\nu$, see
\cite[Eq. 4.11]{Li:97}), and $\sum w_\nu\leq 1$ holds. We will
assume programs to be \emph{binary}, i.e. $p\in\BBB^*$, in
contrast to outputs, which are strings $x\in\XX^*$.

The MDL definitions in Section \ref{secMDL} directly transfer to
this setup. All our results (Theorems
\ref{Theorem:RhoConvergesIMS} - \ref{thSMDLBound}) therefore apply
to $\rrho=\rrho_{[\MM]}$ if the true distribution $\mu$ is a
measure, which is not very restrictive. Then $\mu$ is necessarily
computable. Also, Theorem \ref{Theorem:Solomonoff} implies
Solomonoff's important {\it universal induction} theorem: $\xi$
converges to the true distribution i.m.s., if the latter is
computable. Note that the Bayes mixture $\xi$ is within a
multiplicative constant of the {\it Solomonoff-Levin prior}
$M(x)$, which is the algorithmic probability that $U$ produces an
output starting with
$x$ if its input is random.

In addition to $\MM$, we also consider the set of all recursive
measures $\tilde\MM$ together with the same canonical weights, and
the mixture $\tilde \xi(x)=\sum_{\nu\in\tilde\MM}w_\nu\nu(x)$.
Likewise, define $\tilde\rrho=\rrho_{[\tilde\MM]}$. Then we
obviously have $\tilde\rrho(x)\leq\tilde \xi(x)\leq\xi(x)$ and
$\rrho(x)\leq\xi(x)$ for all $x\in\XX^*$.
It is even immediate that $\xi(x)\leqm\rrho(x)$ since $\xi\in\MM$.
Here, by $f\leqm g$ we mean $f\leq g\cdot O(1)$, ``$\geqm$" and
``$\eqm$" are defined analogously.

Moreover, for any string $x\in\XX^*$, there is also
a {\it universal one-part MDL estimator} $m(x)=2^{-\Km(x)}$
derived from the monotone complexity
$\Km(x)=\min\{\l(p):U(p)=x*\}$. (I.e. the monotone complexity is
the length of the shortest program such that $U$'s output starts
with $x$.) The minimal program $p$ defines a measure
$\nu$ with $\nu(x)=1$ and $w_\nu\geq 2^{-\l(p)}\cdot O(1)$ (recall that
programs are binary). Therefore, $m(x)\leqm\tilde\rrho(x)$ for all
$x\in\XX^*$. Together with the following proposition, we thus
obtain
\beq \label{Eq:ComplexityRelations}
  m(x)\ \eqm\ \tilde\rrho(x)\ \leqm \ \tilde\xi(x) \ \leqm\ \rrho(x)\ \eqm\ \xi(x)
  \for_all x\in\XX^*.
\eeq

\begin{Prop}
We have $\tilde \rrho(x) \leqm m(x)$ for all $x\in\XX^*$.
\end{Prop}

\begin{Proof} (Sketch only.)
It is not hard to show that given a string $x\in\XX^*$ and a
recursive measure
$\nu$ (which in particular may be the MDL descriptor
$\nu^*(x)$) it is possible to specify a program $p$ of length at
most $-\lb w_\nu-\lb\nu(x)+c$ that outputs a string starting with
$x$, where constant $c$ is independent of $x$ and $\nu$. This is
done via arithmetic encoding.
Alternatively, it is also possible to prove the proposition
indirectly using \cite[Th.4.5.4]{Li:97}. This implies that
$m(x)\geqm w_\nu\nu(x)$ for all $x\in\XX^*$ and all recursive
measures $\nu$. Then, also $m(x)\geqm \max\{w_\nu\nu(x)\}$ holds.
\end{Proof}

On the other hand, we know from \cite{Gacs:83} that
$m\neqm\xi$. Therefore, at least one of the
two inequalities in (\ref{Eq:ComplexityRelations}) must be proper.

\begin{Problem}
Which of the inequalities $\tilde \rrho\leqm\tilde \xi$ and
$\tilde \xi\leqm\rrho$ is proper (or are both)?
\end{Problem}

Equation (\ref{Eq:ComplexityRelations}) also has an easy
consequence in terms of randomness criteria.

\begin{Prop}
A sequence $x\ltinf\in\XX^\infty$ is Martin-L{\"o}f random with
respect to some computable measure $\mu$ iff for any
$f\in\{\m,\tilde
\rrho,\tilde\xi,r,\M\}$ there is a constant $C>0$ such that
$f(x\leqn)\leq C\mu(x\leqn)$ for all
$n\in\NNN$ holds.
\end{Prop}

\begin{Proof}
It is a standard result that if $x\ltinf$ is random then
$\M(x\leqn)\leq C\mu(x\leqn)$ for some $C$ \cite[Th.3]{Levin:73random}.
Then by (\ref{Eq:ComplexityRelations}),
$f(x\leqn)\leqm\mu(x\leqn)$ for all $f$. Conversely, if
$f(x\leqn)\leqm\mu(x\leqn)$ for some $f$, then there is $C$ such
that $\m(x\leqn)\leq C\mu(x\leqn)$.
This implies $\mu$-randomness of $x\ltinf$
(\cite[Th.2]{Levin:73random} or \cite[p295]{Li:97}).
\end{Proof}

Interestingly, these randomness criteria partly depend on the
weights. The criteria for $\tilde\xi$ and $\tilde\rrho$ are not
equivalent any more if weights other than the canonical weights
are used, as the following example will show. In contrast, for
$\xi$ and
$\rrho$ there is no weight dependency as long as the weights are
strictly greater than zero, since $\xi\in\MM$.

\begin{Example}
There are other randomness criteria than Martin-L{\"o}f
randomness, e.g.\ rec-randomness. A rec-random sequence $x\ltinf$
(with respect to the uniform distribution) satisfies
$\nu(x\leqn)\leq c(\nu) 2^{-n}$ for each computable measure $\nu$
and for all $n$. It is obvious that Martin-L{\"o}f random
sequences are also rec-random. The converse does not hold, there
are sequences
$x\ltinf$ that are rec-random but not Martin-L{\"o}f random, as
shown e.g.\ in \cite{Schnorr:71,Wang:96}.

Let $x\ltinf$ be such a sequence, i.e.\ $\nu(x\leqn)\leq c(\nu)
2^{-n}$ for all computable measures $\nu$ and for all $n$, but
where $x\ltinf$ is not Martin-L{\"o}f random. Let
$\nu_1,\nu_2,\ldots$ be a (non-effective) enumeration of all
computable measures. Define $w'_i=2^{-i}c(\nu_i)^{-1}$. Then
\beqn
\tilde\M'(x\leqn)=
\sum_{i=1}^\infty w'_i\nu_i(x\leqn)\leq\sum_{i=1}^\infty
2^{-i}c(\nu_i)^{-1}c(\nu_i)2^{-n}=2^{-n}\for_all n,
\eeqn
i.e.\ $x\ltinf$ is $\tilde\M'$-random. Thus, $x\ltinf$ is also
$\tilde r'$-random with $\tilde r'=\max_i\{w'_i\nu_i\}$.
\end{Example}

\section{Conclusions}\label{secDC}

We have proven convergence theorems for MDL prediction for
arbitrary countable classes of semimeasures, the only requirement
being that the true distribution $\mu$ is a measure. Our results
hold for both static and dynamic MDL and provide a statement about
convergence speed in mean sum. This also yields both on-sequence
and off-sequence assertions. Our results are to our knowledge the
strongest available for the discrete case.

Compared to the bound for Solomonoff prediction in Theorem
\ref{Theorem:Solomonoff}, the error bounds for MDL are
exponentially worse, namely $w_\mu^{-1}$ instead of $\ln
w_\mu^{-1}$. Our bounds are sharp in general, as Example
\ref{Ex:LowerBound} shows. There are even classes of Bernoulli
distributions where the exponential bound is sharp
\cite{Poland:04mdlspeed}.

In the case of continuously parameterized model classes, finite
error bounds do not hold \cite{Barron:91,Barron:98}, but the error
grows slowly as $\ln t$. Under additional assumptions (i.i.d. for
instance) and with a reasonable prior, one can prove similar
behavior of MDL and Bayes mixture predictions \cite{Rissanen:96}.
In this sense, MDL converges as fast as Bayes mixture, and this is
even true for the ``slow" Bernoulli example presented in
\cite{Poland:04mdlspeed}. However in Example \ref{Ex:LowerBound},
the error grows as $t$, which shows that the Bayes mixture may be
superior to MDL in general.

\begin{small}
\newcommand{\etalchar}[1]{$^{#1}$}

\end{small}

\end{document}